\documentclass[11pt,a4paper]{article}
\usepackage[hyperref]{acl2021}
\usepackage{times}
\usepackage{latexsym}
\usepackage{graphicx}
\usepackage{adjustbox}
\usepackage{newfloat}
\usepackage{listings}
\usepackage{amsmath}
\usepackage{amsfonts}
\usepackage{amssymb}
\usepackage{makecell}
\usepackage{multirow}
\usepackage{xcolor}
\usepackage{textcomp}
\usepackage{longtable}

\usepackage{microtype}
\usepackage{listings}
\usepackage{xcolor}

\definecolor{codegreen}{rgb}{0,0.6,0}
\definecolor{codegray}{rgb}{0.5,0.5,0.5}
\definecolor{codepurple}{rgb}{0.58,0,0.82}
\definecolor{backcolour}{rgb}{0.95,0.95,0.92}

\lstdefinestyle{mystyle}{
    backgroundcolor=\color{backcolour},   
    commentstyle=\color{codegreen},
    keywordstyle=\color{magenta},
    numberstyle=\tiny\color{codegray},
    stringstyle=\color{codepurple},
    basicstyle=\ttfamily\footnotesize,
    breakatwhitespace=false,         
    breaklines=true,                 
    captionpos=b,                    
    keepspaces=true,                 
    numbers=left,                    
    numbersep=5pt,                  
    showspaces=false,                
    showstringspaces=false,
    showtabs=false,                  
    tabsize=2
}

\aclfinalcopy % Uncomment this line for the final submission

\title{LDKP: A Dataset for Identifying Keyphrases from Long Scientific Documents \thanks{Emails ids of the corresponding authors are, Debanjan.Mahata@moodys.com, yamank@iiitd.ac.in, rajivratn@iiitd.ac.in. Debanjan Mahata participated in this work as an Adjunct Faculty at IIIT-Delhi.}}

\author{Debanjan Mahata \\
  \small{Moody's Analytics} \\ \And
  Navneet Agarwal \\
  \small{IIIT-Delhi} \\\And
  Dibya Gautam \\
  \small{IIIT-Delhi} \\\AND
  Amardeep Kumar \\
  \small{IIT-Dhanbad} \\\And
    Swapnil Parekh \\
  \small{NYU} \\\And
  Yaman Kumar Singla \\
  \small{Adobe Media Data Science Research} \\\AND
    Anish Acharya \\
  \small{UT Austin} \\\And
    Rajiv Ratn Shah \\
  \small{IIIT-Delhi} \\}

\date{}

\begin{document}
\maketitle
\begin{abstract}
Identifying keyphrases (KPs) from text documents is a fundamental task in natural language processing and information retrieval. Vast majority of the benchmark datasets for this task are from the scientific domain containing only the document title and abstract information. This limits keyphrase extraction (KPE) and keyphrase generation (KPG) algorithms to identify keyphrases from human-written summaries that are often very short ($\approx$ 8 sentences). This presents three challenges for real-world applications: human-written summaries are unavailable for most documents, the documents are almost always long, and a high percentage of KPs are \textit{directly} found beyond the limited context of title and abstract. 
%Further, none of the datasets surveyed contains metadata such as author, time, publication venue and year, and the field of study. As shown by previous studies, these information fields impact keyphrase extraction. 
Therefore, we release two extensive corpora mapping KPs of $\approx 1.3M$ and $\approx 100K$ scientific articles with their fully extracted text and additional metadata including publication venue, year, author, field of study, and citations for facilitating research on this real-world problem. 
% Further, we provide an extensive analysis of the released dataset and benchmark the performance of several unsupervised and supervised models for KPE. We show that it is indeed challenging for the popular methods to perform well on longer text and it is necessary for the community to start thinking about developing new methods that can process full-length documents and identify keyphrases from them.

%We observe that processing full-length documents by the recent transformer-based longformer model gives the best results.
\end{abstract}

%%%Intro%%%%%%%%%%%%%%%%%%%%%%%%%%%%%%%%%%%%%
\section{Introduction and Background}
\label{sec:intro}
Identifying keyphrases (KPs) is a form of extreme summarization, where given an input document, the task is to find a set of \textbf{representative} phrases that can effectively summarize it. Over the last decade, we have seen an exponential increase in the velocity at which unstructured text is produced on the web, with the vast majority of them untagged or poorly tagged. KPs provide an effective way to search, summarize, tag, and manage these documents. Identifying KPs have proved to be useful as preprocessing, pre-training \cite{kulkarni2021learning}, or supplementary tasks in other tasks such as search \cite{sanyal2019enhancing, gutwin1999improving, song2006keyphrase}, recommendation systems \cite{augenstein-etal-2017-semeval}, advertising \cite{yih2006finding}, summarization \cite{qazvinian2010citation}, opinion mining \cite{berend2011opinion} to name a few. This has motivated researchers to explore machine learning algorithms for automatically mapping documents to a set of keyphrases commonly referred as the \textit{keyphrase extraction} (KPE) task \cite{kim-etal-2010-semeval, augenstein-etal-2017-semeval}.  

\begin{table*}[!h]
%\adjustbox{width=\textwidth}{
\centering
\scalebox{1.0}{
\begin{tabular} {lllllll}
% {p{0.7\linewidth} p{0.15\linewidth} p{0.05\linewidth}}
\hline
\thead{Dataset}     & \thead{Size} & \thead{Long\\Doc} &  \thead{Avg\\\# Sentences} & \thead{Avg\\\# Words} & \thead{Present\\KPs} & \thead{Absent\\KPs}\\ 
\hline
SemEval2017 \cite{augenstein-etal-2017-semeval} & {\color[HTML]{000000}0.5k}  & {\color[HTML]{000000}$\times$}    & 7.36 &	176.13 & 42.01\% & 57.69\%\\
KDD \cite{KDD-Caragea2014CitationEnhancedKE}        & {\color[HTML]{000000}0.75k}  & {\color[HTML]{000000}$\times$}  & 8.05 &	188.43& 45.99\% & 54.01\%\\
Inspec \cite{inspec}     & {\color[HTML]{000000}2k} & {\color[HTML]{000000}$\times$}  & 5.45 &	130.57 & 55.69\% & 44.31\%\\
KP20k \cite{rui_meng}       & 568k       & {\color[HTML]{000000}$\times$}  & 7.42	& 188.47 & 57.4\% & 42.6\%\\
OAGKx \cite{oagkx}       & 22M   & {\color[HTML]{000000}$\times$}   & 8.87 &	228.50 & 52.7\% & 47.3\% \\ 

\hline
NUS \cite{10.1007/978-3-540-77094-7_41}  & {\color[HTML]{000000}0.21k}  & \checkmark  & 375.93 &	7644.43 & 67.75\%  & 32.25\%\\
SemEval2010 \cite{kim-etal-2010-semeval} & {\color[HTML]{000000}0.24k}  & \checkmark & 319.32 & 7434.52 & 42.01\% & 57.99\%\\
Krapivin \cite{krapivin-2010}    & {\color[HTML]{000000}2.3k} & \checkmark   & 370.48	& 8420.76 & 44.74\% & 52.26\%\\
\hline
\textbf{LDKP3K} (S2ORC $\leftarrow$ KP20K)      & \textbf{100k}   & \checkmark & 280.67 & 6027.10 & 76.11\% & 23.89\%\\
\textbf{LDKP10K} (S2ORC $\leftarrow$ OAGKx)    & \textbf{1.3M}  & \checkmark & 194.76 & 4384.58 & 63.65\% & 36.35\%\\ 
\hline
\end{tabular}}
%}
\caption{Characteristics of the proposed datasets compared to the existing datasets.}
\label{tab:new_datasets_benefits}
\end{table*}

Various algorithms have been proposed over time to solve the problem of identifying keyphrases from text documents that can primarily be categorized into supervised and unsupervised approaches \cite{papagiannopoulou2020review}. Majority of these approaches take an abstract (\textit{a summary}) of a text document as the input and produce keyphrases as output. However, in real world industrial applications in different domains such as advertising \cite{hussain2017automatic}, search and indexing, finance \cite{gupta2020comprehensive}, law \cite{bhargava2017catchphrase}, and many other real-world use cases, document summaries are not readily available. Moreover, most of the documents encountered in these applications are greater than 8 sentences (the average length of abstracts in KP datasets, see Table~\ref{tab:new_datasets_benefits}). We also find that a significant percentage of keyphrases ($>$18\%) are \textit{directly} found beyond the limited context of a document's title and abstract/summary. These constraints limit the potential of currently developed KPE and KPG algorithms to only theoretical pursuits.
% We also find that a significant percentage of keyphrases ($>$18\%) are \textit{directly} found beyond the limited context of a document's title and abstract/summary (see Table~\ref{tab:new_datasets_benefits}).  

% To address this, a few previous studies had proposed to study long-document keyphrase extraction \cite{kim-etal-2010-semeval,krapivin-2010,10.1007/978-3-540-77094-7_41}. However, the number of long documents available are too few ($<$2.3k) for modern training data-hungry deep learning algorithms. To fill this gap, we propose \textbf{l}ong-\textbf{d}ocument \textbf{k}ey\textbf{p}hrase (\textbf{LDKP}) datasets, and the associated task of LDKP extraction. 

% \paragraph{Keyphrase Extraction:} XXX. 
Many previous studies have pointed out the constraints imposed on KPE algorithms due to the short inputs and artificial nature of available datasets \cite{nguyen2010wingnus,hasan2014automatic,cano2019keyphrase,gallina2020large,kontoulis2021keyphrase}. In particular, \citet{cano2019keyphrase} while explaining the limitations of their proposed algorithms, note that the title and the abstract may not carry sufficient topical information about the article, even when joined together. While most datasets in the domain of KPE consist of titles and abstracts \cite{oagkx}, there have been some attempts at providing long document KP datasets as well (Table~\ref{tab:new_datasets_benefits}). \citet{krapivin-2010} released 2,000 full-length scientific papers from the computer science domain. \citet{kim-etal-2010-semeval} in a SemEval-2010 challenge released a dataset containing 244 full scientific articles along with their author and reader assigned keyphrases. \citet{10.1007/978-3-540-77094-7_41} released 211 full-length scientific documents with multiple annotated keyphrases. All of these datasets were released more than a decade ago and were more suitable for machine-learning models available back then. With today's deep learning paradigms like un/semi-supervised learning requiring Wikipedia sized corpora ($>$6M articles), it becomes imperative to update the KPE and KPG tasks with similar sized corpus. % that has keyphrases assigned by both the authors and the readers.
% % So we are introducing two new datasets containing full documents for keyphrase extraction and generation task and providing supervised and unsupervised baseline models on these datasets.

In this work, we develop two large datasets (LDKP - Long Document Keyphrase) comprising of 100K and 1.3M documents for identifying keyphrases from full-length scientific articles along with their metadata information such as venue, year of publication, author information, inbound and outbound citations, and citation contexts, among others. We achieve this by mapping the existing KP20K \cite{rui_meng} and OAGKx \cite{oagkx} corpus for KPE and KPG to the documents available in S2ORC dataset \cite{lo-etal-2020-s2orc}. We make the dataset publicly available on Huggingface hub (Section \ref{sec::dataset-usage}) in order to facilitate research on identifying keyphrases from long documents. 
% We perform extensive experiments using a number of learning paradigms such as graph-based, statistical, and supervised approaches and identify the challenges they face in the context of long inputs. We show that the best results are obtained by modelling on longer contexts using Longformer \cite{longformer} and treating KPE as a sequence tagging task \cite{sahrawat2020keyphrase}. 
We hope that researchers working in this area would acknowledge the shortcomings of the popularly used datasets and methods in KPE and KPG and devise exciting new approaches for overcoming the challenges related to identifying keyphrases from long documents and contexts beyond summaries. This would make the algorithms more useful in practical real-world settings.

%%%Dataset%%%%%%%%%%%%%%%%%%%%%%%%%%%%%%%%%%%%%
\section{Dataset}
\label{sec:dataset}
We propose two datasets resulting from the mapping of S2ORC with KP20K and OAGKx corpus, respectively. \citet{lo-etal-2020-s2orc} released S2ORC as a huge corpus of 8.1M scientific documents. While it has full text and metadata (see Table \ref{tab:metadata_table}) the corpus does not contain keyphrases. We took this as an opportunity to create a new corpus for identifying keyphrases from full-length scientific articles. Therefore, we took the KP20K and OAGKx scientific corpus for which keyphrases were already available and mapped them to their corresponding documents in S2ORC. This is the first time in the keyphrase community that such a large number of full-length documents with comprehensive metadata information have been made publicly available for academic use.

We release two datasets LDKP3K and LDKP10K corresponding to KP20K and OAGKx, respectively. The first corpus consists of $\approx$ \textbf{100K} keyphrase tagged long documents obtained by mapping KP20K to S2ORC. The KP20K corpus mainly contains title, abstract and keyphrases for computer science research articles from online digital libraries like ACM Digital Library, ScienceDirect, and Wiley. Using S2ORC documents, we increase the average length of the documents in KP20K from 7.42 sentences to 280.67 sentences, thereby also increasing the percentage of present keyphrases in the input text by 18.7\%. 

The second corpus corresponding to OAGKx consists of \textbf{1.3M} full scientific articles from various domains with their corresponding keyphrases collected from academic graph \cite{sinha2015overview,tang2008arnetminer}.  The resulting corpus contains 194.7 sentences (up from 8.87 sentences) on an average with 63.65\% present keyphrases (up from 52.7\%). Since both datasets consist of a large number of documents, we present three versions of each dataset with the training data split into \textit{small}, \textit{medium} and \textit{large} sizes, as given in Table \ref{tab:new_datasets}. This was done in order to provide an opportunity even to researchers and practitioners with scarcity of computing resources to evaluate the performance of their methods on a smaller dataset that can be trained in free platforms like Google Colab\footnote{https://colab.research.google.com/}.

\begin{figure}[htp]
    \includegraphics[width=8cm]{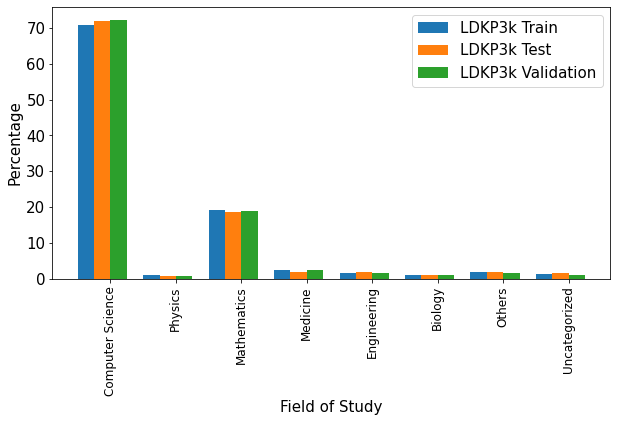}
    \caption{Field of Studies distribution for train, test and validation  split of LDKP3K dataset.}
    \label{figure:ldkp3k_fos}
\end{figure}

\begin{figure}[htp]
    \includegraphics[width=8cm]{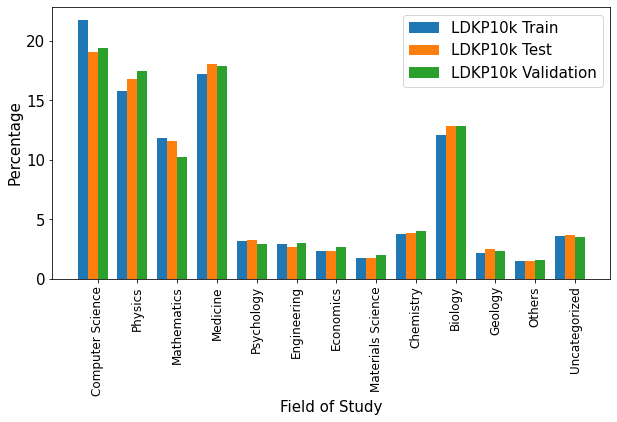}
    \caption{Field of Studies distribution for train, test and validation  split of LDKP10K dataset.}
    \label{figure:ldkp10k_fos}
\end{figure}

\subsection{Dataset Preparation}
In the absence of any unique identifier shared across datasets, we used paper title to map documents in S2ORC with  KP20K/OAGKx. This had its own set of challenges. For example, some papers in KP20K and OAGKx had unigram titles like ``Editorial" or ``Preface". Multiple papers can be found with the same title. We ignored all the papers with unigram and bigram titles. We resolved the title conflicts through manual verification.

We also found out that some of the keyphrases in OAGKx and KP20K datasets were parsed incorrectly. Keyphrases that contain delimiters such as \textit{comma} (which is also used as a separator for keyphrase list) have been broken down into two or more keyphrases, \textit{e.g.}, the keyphrase `\textit{2,4- dichlorophenoxyacetic acid}' has been broken down into [`2', `4- dichlorophenoxyacetic acid']. In some cases, the publication year, page number, DOI, \textit{e.g.}, \textit{1999:14:555-558}, were inaccurately added to the list of keyphrases. To solve this, we filtered out all the keyphrases that did not have any alphabetical characters in them. 

Next, in order to facilitate the usage of particular sections in KPE algorithms, we standardized the section names across all the papers. The section names varied across different papers in the S2ORC dataset. For example, some papers have a section named \textit{``Introduction"} while others have it as \textit{``1.Introduction", ``I. Introduction", ``I Introduction" etc}. To deal with this problem, we replaced the unique section names with a common generic section name, like \textit{``introduction"}, across all the papers. We did this for common sections including introduction, related work, conclusion, methodology, results, and analysis.

% \textbf{\textit{Describe how the train/dev/test} splits were decided. Look at the distribution of the topics in train/dev/test. Look at the distribution of all the aspects like length etc in train/dev/test - avg number of sentences, words, etc}

The proposed dataset LDKP3k and LDKP10k are further divided into train, test and validation splits as shown in Table-\ref{tab:new_datasets}. For LDKP3k, these splits are based on the splits that was present in the original KP20k dataset. For LDKP10k, we resorted to random sampling method to create these splits since OAGKX, the keyphrase dataset corresponding to LDKP10k, wasn't originally divided into train, test and validation. Figures \ref{figure:ldkp3k_fos} and \ref{figure:ldkp10k_fos} show that the splits for both LDKP3k and LDKP10k are of adequate quality because there is a good distribution of papers in terms of field of study across all the splits.

% The number of validation and test samples are 3K for \textbf{LDKP3k} and 10k for \textbf{LDKP10k}.

% We have created two splits of this new dataset called \textbf{LDKP3k} and \textbf{LDKP10k} which are inspired from KP20k and OAGkx respectively.

\begin{table}[]
\begin{tabular}{|l|l|l|l|}
\hline
\multicolumn{2}{|l|}{\textbf{Dataset}}            & \textbf{LDKP3K} & \textbf{LDKP10K} \\ \hline
\multirow{3}{*}{\textbf{Train}} & \textbf{Small}  & 20,000          & 20,000          \\ \cline{2-4} 
                                & \textbf{Medium} & 50,000          & 50,000          \\ \cline{2-4} 
                                & \textbf{Large}  & 90,019          & 1,296,613        \\ \hline
\multicolumn{2}{|l|}{\textbf{Test}}               & 3,413           & 10,000          \\ \hline
\multicolumn{2}{|l|}{\textbf{Validation}}         & 3,339           & 10,000          \\ \hline
\end{tabular}
\caption{LDKP datasets with their train, validation and test dataset distributions.}
\label{tab:new_datasets}
%\vspace{-3 mm}
\end{table}
% \begin{figure}[hbt!]
%     \centering
%     \centering
%     \includegraphics[scale=0.5]{plots/dataset/oagkx_freq.png}
%     \caption{Keyphrase Frequency distibution for \textbf{LDKP10k}}
%     \label{fig:freq_kp_oagkx}
% \end{figure}

% \begin{figure}
%     \centering
%     \centering
%     \includegraphics[scale=0.5]{plots/dataset/kp20k_freq.png.png}
%     \caption{Keyphrase Frequency distribution for \textbf{LDKP3k}}
%     \label{fig:freq_kp_kp20k}
% \end{figure}

% \label{table:metadata_table}
% \input{metadata_table}

\begin{table}[]
\adjustbox{width=\columnwidth}{
\begin{tabular}{lll}
\hline
\textbf{\thead{Paper \\ details}} & \textbf{\thead{Paper \\ Identifier}}   & \textbf{\thead{Citations and \\ References}} \\ \hline
Paper ID               & ArXiv ID                    & Outbound Citations                \\
Title                  & ACL ID                      & Inbound Citations                 \\
Authors                & PMC ID                      & Bibliography                      \\
Year                   & PubMed ID                   & References                        \\
Venue                  & MAG ID                       &                                   \\
Journal                & DOI &                                   \\
Field of Study         & S2 URL   &      
\\\hline
\end{tabular}
}
\caption{Information available in the metadata of each scientific paper in LDKP corpus.}
\label{tab:metadata_table}
\end{table}

\section{Dataset Usage}
\label{sec::dataset-usage}
% \begin{lstlisting}[language=Python]
% from datasets import load_dataset

% # get small dataset
% dataset = load_dataset("midas/ldkp3k", "small")
% \end{lstlisting}

Please refer to the Huggingface hub pages for LDKP3k and LDKP10k for a detailed information about downloading and using the dataset.

\begin{enumerate}
    \item LDKP3K - \url{https://huggingface.co/datasets/midas/ldkp3k}
    \item LDKP10K - \url{https://huggingface.co/datasets/midas/ldkp10k}
\end{enumerate}

\section{Conclusion}
In this work, we identified the shortage of corpus comprising of long documents for training and evaluating keyphrase extraction and generation models. We created two very large corpus - LDKP3K and LDKP10K comprising of $\approx$ 100K and $\approx$ 1.3M documents and make it publicly available. We hope this would encourage the researchers to innovate and propose new models capable of identifying high quality keyphrases from long multi-page documents.

% %%%Experiments%%%%%%%%%%%%%%%%%%%%%%%%%%%%%%%%%%%%%
% \input{acl-ijcnlp2021-templates/experiments}

% %%%Interpretability%%%%%%%%%%%%%%%%%%%%%%%%%%%%%%%%%%%%%
% \input{acl-ijcnlp2021-templates/interpretability}

\bibliographystyle{acl_natbib}
\bibliography{anthology,acl2021}

%\appendix

\end{document}